\documentclass[11pt,twoside]{article} 

\usepackage{amsmath,amsfonts,bm, dsfont}

\usepackage{changes}
\definechangesauthor[name=Khang Le, color=red]{KL}

\def\XX{\mathbf{X}}

\def\CC{\mathbf{C}}

\def\aA{\mathbf{a}}
\def\bB{\mathbf{b}}
\def\pP{\mathbf{p}}

\def\rR{\mathbf{r}}

\def\xX{\mathbf{x}}

\newcommand{\br}{\mathbb{R}}

\newcommand{\ba}{\begin{array}}
\newcommand{\ea}{\end{array}}

\newcommand{\bigO}{\mathcal{O}}
\newcommand{\bigOtil}{\widetilde{\mathcal{O}}}

\newcommand{\OT}{\mathbf{OT}}
\newcommand{\POT}{\mathbf{POT}}

\newcommand{\zeros}{\mathbf{0}}
\newcommand{\ones}{\mathbf{1}}

\newcommand{\onorm}[1]{\|#1\|_{1}}

\usepackage{enumitem}

\usepackage[utf8]{inputenc} 
\usepackage[T1]{fontenc}    
\usepackage{booktabs}       
\usepackage{amsfonts}       
\usepackage{nicefrac}       
\usepackage{microtype}      

\usepackage{subfigure}
\usepackage{epsf}
\usepackage{epsfig}
\usepackage{fancyhdr}
\usepackage{graphics}
\usepackage{graphicx}
\usepackage{psfrag}
\usepackage{pdfpages}

\usepackage{url}
\usepackage[colorlinks,linkcolor=magenta,citecolor=blue, pagebackref=true,backref=true]{hyperref}
\renewcommand*{\backrefalt}[4]{%
    \ifcase #1 \footnotesize{(Not cited.)}%
    \or        \footnotesize{(Cited on page~#2.)}%
    \else      \footnotesize{(Cited on pages~#2.)}%
    \fi}
\usepackage{amsthm}
\usepackage{amsmath}
\usepackage{amssymb,bbm}
\usepackage{caption}
\usepackage{algorithmic}
\usepackage{algorithm}
\usepackage{multicol}
\usepackage{caption}
\usepackage{color}

\DeclareMathOperator*{\argmin}{arg\,min}

\newtheorem{assumption}{Assumption}
\newtheorem{remark}{Remark}

\newtheorem{lemma}{Lemma}
\newtheorem{theorem}{Theorem}
\newtheorem{proposition}{Proposition}
\newtheorem{definition}{Definition}

\newtheorem{example}{Example}

\usepackage{changes}
\definechangesauthor[name=Huy Nguyen, color=purple]{HN}
\usepackage{natbib}

\usepackage[utf8]{inputenc} 
\usepackage[T1]{fontenc}    
\usepackage{booktabs}       
\usepackage{amsfonts}       
\usepackage{nicefrac}       
\usepackage{microtype}      

\usepackage{subfigure}
\usepackage{epsf}
\usepackage{epsfig}
\usepackage{fancyhdr}
\usepackage{graphics}
\usepackage{graphicx}
\usepackage{psfrag}
\usepackage{fullpage}
\usepackage{pdfpages}

\usepackage{url}
\usepackage[colorlinks,linkcolor=magenta,citecolor=blue, pagebackref=true,backref=true]{hyperref}
\renewcommand*{\backrefalt}[4]{%
    \ifcase #1 \footnotesize{(Not cited.)}%
    \or        \footnotesize{(Cited on page~#2.)}%
    \else      \footnotesize{(Cited on pages~#2.)}%
    \fi}

\usepackage{amsthm}
\usepackage{amsmath}
\usepackage{amssymb,bbm}
\usepackage{caption}
\usepackage{algorithmic}
\usepackage{algorithm}
\usepackage{multicol}
\usepackage{caption}
\usepackage{color}
\begin{document}

%

%

\begin{center}
{\bf{\LARGE{On Multimarginal Partial Optimal Transport: \\ Equivalent Forms and Computational Complexity}}}
  
\vspace*{.2in}
{\large{
\begin{tabular}{ccccc}
Khang Le$^{\star, \dagger}$ & Huy Nguyen$^{\star, \diamond}$ & Khai Nguyen$^{\dagger}$ & Tung Pham$^{\diamond}$  & Nhat Ho$^{\dagger}$\\
\end{tabular}
}}

\vspace*{.2in}

\begin{tabular}{c}
University of Texas, Austin$^{\dagger}$; VinAI Research$^\diamond$\\
\end{tabular}

\vspace*{.2in}

\begin{abstract}
  We study the multimarginal partial optimal transport (POT) problem between $m$ discrete (unbalanced) measures with at most $n$ supports. We first prove that we can obtain two equivalent forms of the multimarginal POT problem in terms of the multimarginal optimal transport problem via novel extensions of cost tensors. The first equivalent form is derived under the assumptions that the total masses of each measure are sufficiently close while the second equivalent form does not require any conditions on these masses but at the price of more sophisticated extended cost tensor. Our proof techniques for obtaining these equivalent forms rely on novel procedures of moving masses in graph theory to push transportation plan into appropriate regions. Finally, based on the equivalent forms, we develop an optimization algorithm, named the ApproxMPOT algorithm, that builds upon the Sinkhorn algorithm for solving the entropic regularized multimarginal optimal transport. We demonstrate that the ApproxMPOT algorithm can approximate the optimal value of multimarginal POT problem with a computational complexity upper bound of the order $\bigOtil(m^3(n+1)^{m}/ \varepsilon^2)$ where $\varepsilon > 0$ stands for the desired tolerance.
\end{abstract}
\let\thefootnote\relax\footnotetext{$\star$ Khang Le and Huy Nguyen
  contributed equally to this work.}
 \end{center}
\section{Introduction}
\label{sec:introduction}
The recent advances in computation of optimal transport (OT)~\citep{Villani-03, Cuturi-2013-Sinkhorn, Lin-2019-Efficient, Dvurechensky-2018-Computational, Altschuler-2017-Near} have brought new applications of optimal transport in machine learning and data science to the fore. Examples of these applications include generative models~\citep{Arjovsky-2017-Wasserstein, tolstikhin2018wasserstein, Gulrajani-2017-Improved, Courty-2017-Optimal}, unsupervised learning~\citep{Ho-ICML-2017, Ho_JMLR}, computer vision~\citep{Solomon_2015, Nguyen_3D}, and other applications~\citep{Rolet-2016-Fast, Peyre-2016-Averaging, Carriere-2017-Sliced}. However, due to the marginal constraints of transportation plans, optimal transport is only defined between balanced measures, namely, measures with equal masses.

When measures are unbalanced, i.e., they can have different masses, there are two popular approaches for defining divergences between these measures. The first approach is unbalanced optimal transport~\citep{Chizat_Unbalanced,Chizat_Scaling_2016}. The main idea of unbalanced optimal transport is to regularize the objective function of optimal transport based on certain divergences between marginal constraints of transportation plan and the masses of the input measures. Despite its favorable computational complexity~\citep{pham2020unbalanced} and practical applications~\citep{Schiebinger_Optimal_2019, frogner2015learning, Janati_Wasserstein_2019, Balaji_Robust}, the optimal transportation plan from unbalanced optimal transport is often non-trivial to interpret in practice. 

The second approach for defining divergence between unbalanced measures is partial optimal transport (POT)~\citep{Caffarelli_POT, Figalli_POT}, which was originally used to analyze partial differential equations. The idea of partial optimal transport is to constrain the total masses that 
will be transported between measures. It requires that the marginals of transport plan to be dominated by the corresponding measure. 
Due to these explicit constraints, the optimal transportation plan obtained from the partial optimal transport is more convenient to interpret than that achieved from the unbalanced optimal transport. As a consequence,  partial optimal transport has begun to be employed in several machine learning applications recently. Examples of these applications include computer graphics~\citep{bonneel2019spot}, graph neural networks~\citep{Sarlin_Graph}, positive-unlabeled learning~\citep{chapel2020partial}, and partial covering~\citep{Kawano_Partial}. To improve the scalability of partial optimal transport in these applications, the tree-sliced version of partial optimal transport has also been recently developed in~\citep{Le_POT}.

In this paper, we put the focus on POT, in particular its natural extension to deal with multiple (more than two) measures, namely multimarginal partial optimal transport (MPOT)~\citep{Kitagawa_MPOT}. Since MPOT has already been studied in previous works (e.g., \citep{Kitagawa_MPOT}), our main goal here is to provide an efficient tool for solving it. Specifically, we are interested in the computational aspect of MPOT in the discrete setting. The current literature offers two possible tools to solve MPOT: linear programming algorithms and the iterative Bregman projection (IBP)~\citep{Benamou-2015-Iterative} . The former comes from the fact that MPOT problem is a linear program (LP) (see \eqref{eq:MPOT_objective}), thus it can be solved by any linear programming algorithm. However, the number of variables and constraints in this case grow exponentially in the number of measures (i.e., $n^m$ variables and $n^m + nm + 1$ constraints for $m$ measures, each having $n$ supports). Thus, the interior point methods for LPs may suffer from worst-case complexity bound $\mathcal{O}(n^{3m})$, while efficient network simplex algorithms used in the two-marginal case is inapplicable (see \citep[Theorem 3.3]{Lin_2020} for the argument why the multimarginal formulation is not a minimum-cost flow problem when $m \ge 3$). On the other hand, the IBP algorithm is difficult to analyze. Inspired by current development in the computational tools~\citep{Lin_2020} for multimarginal optimal transport (MOT)~\citep{Gangbo-1998-MOT, Pass-2015-Multi}, we show that these advances can be used to provide efficient tools to solve MPOT, via novel equivalent forms between MOT and MPOT.

\noindent
\textbf{Contribution.} In this paper, we provide two equivalent forms for MPOT in terms of MOT via novel extensions of cost tensors. The first equivalent form is a natural development of the result in bi-marginal case established in~\citep{chapel2020partial}, but it requires the masses of all measures to be close to each other, which can be quite restrictive in general settings. To account for this limitation, we introduce the second equivalent form, which is more involved but free of additional conditions on the masses. The novel proof techniques for both equivalent forms come from the design of the extended cost tensors and the process of moving masses such that the marginal constraints of the transportation plan are still satisfied. Most of the challenges in our proofs are in the second equivalent form in which we have to design a more sophisticated mass-moving procedure in order to push the masses into $m+1$ distant regions rather than neighbour regions in the first form. Finally, via these equivalent forms, we develop ApproxMPOT algorithm (which is built upon the Sinkhorn algorithm for solving the entropic regularized MOT~\citep{lin2019complexity}) for approximating the solution of the MPOT problem. We prove that this algorithm has a complexity upper bound of the order $\bigOtil(m^3(n+1)^m/ \varepsilon^2)$ for approximating the optimal MPOT cost where $\varepsilon > 0$ is the tolerance.

\noindent
\textbf{Organization.} The paper is organized as follows. In Section~\ref{sec:preliminary}, we provide background on optimal transport and multimarginal partial optimal transport. We then present two equivalent forms of multimarginal partial optimal transport in Section~\ref{Sec:equivalent_forms}. Based on these equivalent forms, we develop ApproxMPOT algorithm for approximating the multimarginal POT and derive a complexity upper bound of that algorithm. We then present the experiments with the ApproxMPOT algorithm in Section~\ref{sec:empirical_study} and conclude the paper in Section~\ref{Sec:conclusion}. Finally, we defer the missing proofs to the Appendices. 

\noindent
\textbf{Notation.} We let $[n]$ stand for the set $\{1, 2, \ldots, n\}$ while $\br^n_+$ stands for the set of all vectors in $\br^n$ with non-negative entries. For a vector $\xX \in \br^n$ and $p\in [1,\infty)$, we denote $\|\xX\|_p$ as its $\ell_p$-norm and $\text{diag}(\xX)$ as the diagonal matrix with $\xX$ on the diagonal. The natural logarithm of a vector $\aA = (a_1,..., a_n) \in \br^n_+$ is denoted $\log \aA = (\log a_1,..., \log a_n)$. $\ones_n$ stands for a vector of length $n$ with all of its components equal to $1$. Let $\Sigma_N$ be the histogram of $N$ bins $\{\pP\in\br^N_+:\sum_{i=1}^Np_i=1\}$ and $\delta$ is the Dirac function. Next, given the dimension $n$ and accuracy $\varepsilon$, the notation $a = \bigO\left(b(n,\varepsilon)\right)$ means that  $a \leq C \cdot b(n, \varepsilon)$ where $C$ is independent of $n$ and $\varepsilon$. Similarly, the notation $a = \bigOtil(b(n, \varepsilon))$ indicates the previous inequality may scale by a logarithmic function of $n$ and $\varepsilon$. The entropy of a matrix $\XX$ is given by $H(\XX)=\sum_{i,j=1}^n-X_{ij}(\log X_{ij}-1)$. In general, we use $\XX$ for transportation matrix/tensor and $X_{(i,j)}$ or $X_{u}$ for $u$ in  $[n+1]^m$ in the multimarginal case  be its corresponding entries.  We also define $\XX_S = \{ X_u: u\in S\}$, where $S$ is a subset of $[n+1]^m$. Lastly, the cost matrix/tensor is denoted by $\mathbf{C}$, where its entries are denoted by $C_v$ for $v\in [n]^m$.

\section{Preliminary}
\label{sec:preliminary}
In this section, we first provide background for optimal transport between two discrete probability measures in Section~\ref{sec:OT}. Then, we present (multimarginal) partial optimal transport between discrete (unbalanced) measures in Sections~\ref{sec:POT} and~\ref{sec:MPOT}.
\subsection{Optimal Transport}
\label{sec:OT}
Assume that $P$ and $Q$ are discrete measures with at most $n \geq 1$ supports such that $P = \sum_{i = 1}^{n} a_{i} \delta_{x_{i}}$ and $Q = \sum_{j = 1}^{n} b_{j} \delta_{y_{j}}$. 
When $P$ and $Q$ have the same mass, i.e., $\|\aA\|_{1} = \|\bB\|_{1}$, the optimal transport (OT) distance between $P$ and $Q$ admits the following form:
\begin{align*}
    \OT(P, Q):=\min_{\XX\in\Pi(\aA,\bB)}\langle \CC,\XX\rangle,
\end{align*}
where $\CC$ is a cost matrix whose entries are distances between the supports of these distributions, $\aA : = (a_{1}, \ldots, a_{n}), \bB : = (b_{1}, \ldots, b_{n})$, and the set of admissible couplings $\Pi(\aA,\bB)$ is given by:
\begin{align}
    \Pi(\aA,\bB) : = \{\XX\in\br^{n\times n}_+:\XX\ones_{n}= \aA, \XX^{\top}\ones_n=\bB\}. \label{eq:OT_formulation}
\end{align}

\subsection{Partial Optimal Transport}
\label{sec:POT}
As indicated in equation~\eqref{eq:OT_formulation}, the OT formulation requires that $P$ and $Q$ have the same mass, i.e., $P$ and $Q$ are balanced measures. To account for the settings when $P$ and $Q$ are unbalanced measures, namely, $\|\aA\|_{1} \neq \|\bB\|_{1}$, we consider in this section an approach named Partial Optimal Transport (POT), which focuses on optimally transporting only a fraction $0\leq s \leq \min \{\|\aA\|_{1}, \|\bB\|_{1}\}$ of the mass. The set of all admissible couplings in that case is $\Pi^s(\aA,\bB) := \{\XX\in\br^{n\times n}_+:\XX\ones_{n} \leq \aA, \XX^{\top}\ones_n\leq\bB, \ones^{\top}_n\XX\ones_{n}=s\}$, and the formulation of the POT distance can be written as 
\begin{align*}
    \POT(P,Q):=\min_{\XX\in\Pi^s(\aA,\bB)}\langle \CC,\XX\rangle.
\end{align*}
At the first sight, the POT problem seems to be more demanding with the appearance of two inequality constraints in $\Pi^s(\aA,\bB)$. However, we can tackle these inequality constraints by adding dummy points $x_{n+1}, y_{n+1}$ and extending the cost matrix as follows~\citep{chapel2020partial}:
\begin{align*}
    \Bar{\CC}=
    \begin{pmatrix}
    \CC & \zeros_n\\
    \zeros^{\top}_n & A
    \end{pmatrix},
\end{align*}
in which $A>0$. We denote $\Bar{\aA}=[\aA,\onorm{\bB}-s]$ and $\Bar{\bB}=[\bB,\onorm{\aA}-s]$. The following result from~\citep{chapel2020partial} demonstrates that solving the POT problem between $P$ and $Q$ is equivalent to solving an OT problem between two measures with mass $\onorm{\aA} + \onorm{\bB}-s$. 
\begin{proposition}[\citep{chapel2020partial}]
\label{prop:POT_OT_relation}
Under the assumption that $A>0$, we have
\begin{align*}
    \min_{\XX\in\Pi^s(\aA,\bB)}\langle \CC,\XX\rangle = \min_{\Bar{\XX}\in\Pi(\Bar{\aA},\Bar{\bB})}\langle \Bar{\CC},\Bar{\XX}\rangle
\end{align*}
and the optimal solution $\XX^*$ of the POT problem is exactly that of the OT problem, denoted by $\Bar{\XX}^*$, deprived from its last row and column.
\end{proposition}
\noindent
The equivalent form in Proposition~\ref{prop:POT_OT_relation} indicates that by using the Sinkhorn algorithm for solving the entropic version of OT problem with cost matrix $\Bar{\CC}$ and the masses $\Bar{\aA}$, $\Bar{\bB}$, the computational complexity of approximating the POT between two measures $P$ and $Q$ will be at the order of $\mathcal{O} \left(\frac{(n + 1)^2}{\varepsilon^2} \right)$ where $\varepsilon > 0$ stands for the desired tolerance.

\subsection{Multimarginal Partial Optimal Transport}
\label{sec:MPOT}
We now consider an extension of the partial optimal transport to multimarginal partial optimal transport (MPOT) when we have more than two measures. Assume that we are given $m$ measures $P_{1}, \ldots, P_{m}$ with at most $n$ supports that have weights $\rR_{1}, \ldots, \rR_{m}$, respectively. The transport cost tensor/matrix denoted by $\mathbf{C} \in \mathbb{R}_{+}^{n^{m}}$ is also known in advance. Given a total mass of $s$ that is needed to be transported,  the multimarginal partial optimal transport wants to find a transport plan $\XX^*$ such that the $i$-th marginals of $\XX^*$ is a sub-measure of $\rR_i$ for all $i\in[m]$ and the total transport cost is minimized. In formula, $\XX^*$ is an optimal solution of the following problem:
\begin{align}
   \textbf{MPOT}(P_{1}, \ldots, P_{m}) := \min_{X \in \Pi^{s}(\rR_{1}, \ldots, \rR_{m})} \langle \CC,\XX\rangle, \label{eq:MPOT_objective} 
\end{align}
where $\Pi^{s}(\rR_{1}, \ldots, \rR_{m})$ is defined as:
\begin{align}
    \label{eq:multi_couplings}
    \Pi^{s}(\rR_{1}, \ldots, \rR_{m}) : = \Big\{\XX\in\br^{n^m}_+
    : c_{k}(\XX) \leq \rR_{k},\forall k \in [m]; \sum_{v \in [n]^m} X_v=s\Big\},
\end{align}
with $\big[c_{k}(\XX)\big]_{j} = \sum_{ i_{\ell}\in [n],\ell\neq k} X_{i_{1},\ldots,i_{k - 1},j,i_{k + 1},\ldots,i_{m}}$. 
\section{Equivalent Forms of Multimarginal Partial Optimal Transport}
\label{Sec:equivalent_forms}
In this section, we first present the challenges of extending the POT-OT equivalence in Proposition \ref{prop:POT_OT_relation} to the multimarginal case. Then, we provide two approaches to link the multimarginal POT problem with its multimarginal OT (MOT) counterpart by cost tensor expansions. Subsequently, we will look into an algorithm to approximate the solution of multimarginal POT problem, which is inspired by two equivalent forms.

Our proofs for the equivalent forms of multimarginal POT mostly play with ``layers'' of the (cost or plan) tensors, which is defined as follow: for a subset $S$ of $\{1, 2, \dots, m\}$, we denote by $T_S$ the subset of tensor indices
\begin{align*}
T_S := \big\{ (i_1,\ldots,i_m)\in [n+1]^m: 
              i_{\ell} = n+1 \Leftrightarrow \ell\in S \big\}.
\end{align*}
For example, when $n=2$ and $m = 3$ and $S = \{1,2 \}$, then $T_S = \{(3,3,1), (3,3,2) \}$. For all subsets $S$ of the same cardinality $k$, the union of $T_S$ (i.e.,  $\cup_{|S| = k} T_S$) forms a layer of order $k$ (namely the $k$-th layer), and if $|S| = k$ then $T_S$ is considered as a sublayer of this union. Thus, there are $m+1$ layers in the $m$-dimensional hypercube. 

\subsection{The Three-Marginal Partial Optimal Transport}
To begin with, we will investigate the simplest case of multimarginal POT, i.e., when $m = 3$. We will examine the structure of this problem under this simple case, and show the reason why we have two approaches for deriving equivalent forms for the general case.

In Figure \ref{figure:illustration_equivalence1}, the red square (or cube) denotes the cost matrix (or tensor) $\mathbf{C}$, corresponding to the case $m = 2$ (or $m = 3$). When $m=2$, by adding dummy points, the red square is extended by adding two green segments and a blue point corresponding to $C_{(n+1,n+1)}$.~\citep{chapel2020partial} then show that the supports (i.e., positive entries) of the optimal transport plan lie on the red and green regions, with the total mass of the red region being equal to $s$, by solving two linear equations of the marginals. Given the uniform cost on the green parts, the OT problem reduces to minimizing 
its objective function in the red square. Thus, solving the bi-marginal POT is equivalent to solve a OT problem.  For $m=3$, there are two ways to extend the cost tensor to form two different systems of equations in order to prove that the total mass of the red part is equal to $s$.

The first approach is to build a cost tensor such that the supports of the transport plan lie only on the red and green regions in Figure \ref{figure:illustration_equivalence1}, and the second approach is to build a cost tensor such that the supports lie only on the red and blue regions. In the language of ``layers'', in the case $m = 3$, the green part and the blue part can be explicitly defined as $\{v \in T_S : S \subset \{1, 2, 3\}, |S| = 1\}$ and $\{v \in T_S : S \subset \{1, 2, 3\}, |S| = 2\}$, respectively.

The key idea of the equivalent forms is how to design mass-moving procedures inside the hyper-cube that keep the marginal constraints and decrease the objective function. After applying the procedure until we could not carry on further, the mass concentrates on the certain parts of the transport plan so that  we can derive a system of linear equations to show that the total mass in the red region  equals to $s$. Together with  either the mass or the cost of the same layer of order $1,2,\ldots,m$ must be zero, it deduces  the equivalence between the multimarginal POT and the multimarginal OT problems. Because the moving-mass procedure transports mass between neighbouring layers/regions, the closer between non-zero parts of the transport plan are, the easier to design plan to push mass into those parts is. From this point, we could see that the first extension is the more natural extension of the bi-marginal case, when the non-zero parts (red and green) of the optimal transport plan are close to each other under the Hamming distance. That leads to a simple design of the cost tensor. The second extension requires  novel technique to deal with, since the red (non-zero) parts of the optimal transport plan are far from the blue parts. That requires a more sophisticated  mass moving procedure as well as the more complicated structure of the weights in the cost tensor.

\subsection{The First Equivalent Form}
Inspired by the extension technique in the bi-marginal case, we introduce the first equivalent form for the multimarginal POT problem. However, this equivalence comes at a cost of introducing additional conditions on the marginals that read
\begin{align}
    \label{asummption:mpot_equivalence_1}
    \Sigma_r := \sum_{i=1}^m \|\rR_i\|_1 \geq (m-1) \|\rR_k\|_1 + s, ~ \forall k \in [m].
\end{align}
These conditions are necessary for the extended marginals to be non-negative (see Theorem \ref{theorem:mpot_equivalence_1}). Note that these system of inequalities always hold for $m = 2$, and still hold for $m \ge 3$ in several cases, for example, the total masses of each measure are the same (i.e., $\|\rR_i\|_1 = \|\rR_j\|_1$ for all $i, j \in [m]$). We now state the first equivalent form under the conditions~\eqref{asummption:mpot_equivalence_1}.

\begin{theorem}
\label{theorem:mpot_equivalence_1}
Assume that the conditions \eqref{asummption:mpot_equivalence_1} are met. Let
\begin{align*}
    \bar{\rR}_k^{(1)}=\Big[\rR_k, \frac{1}{m - 1} \Sigma_r - \onorm{\rR_k} - \frac{1}{m - 1} s\Big]
\end{align*}
be extended marginals and  $\bar{\CC}^{(1)} = \big[\bar{C}_v^{(1)}: v\in [n+1]^m \big]$, where
\begin{align*}
    \bar{C}_{v}^{(1)}:=
    \begin{cases}
        C_{v}& \quad v \in T_{\varnothing} \\
        A_i &\quad v \in T_S, |S| = i \quad \forall i \in [m]
    \end{cases}
\end{align*}
be an extended cost tensor in which $0 = A_1 < A_2 < \ldots < A_{m}$. Then, we find that
\begin{align}
    \label{eq:mpot_mot_1}
    \min_{\XX \in \Pi^{s}(\rR_{1}, \ldots, \rR_{m})} \langle \CC,\XX\rangle = \min_{\bar{\XX} \in \Pi(\bar{\rR}_{1}^{(1)}, \ldots, \bar{\rR}_{m}^{(1)})} \langle \bar{\CC}^{(1)},\bar{\XX}\rangle.
\end{align}
Moreover, assume that $\Bar{\XX}^*$ is a minimizer of the multimarginal OT problem, then $\XX^*:=\Bar{\XX}^*_{[n]^m}$ is an optimal solution of the multimarginal POT problem.
\end{theorem}
\begin{example}
\label{example:first_equivalence}
To illustrate the result of Theorem~\ref{theorem:mpot_equivalence_1}, we assume that $m = 3$, i.e., 3-marginal setting, and $\|r_{1}\|_{1} = \|r_{2}\|_{1} = \|r_{3}\|_1 = 1$. Under this setting, the condition~\eqref{asummption:mpot_equivalence_1} is satisfied. For simplicity, we set $n=2$, meaning each marginal has two support points. Based on Theorem~\ref{theorem:mpot_equivalence_1}, the extended cost tensor $\bar{\CC}^{(1)}$ takes the following form: $\bar{C}^{(1)}_{u_1} = 0,\bar{C}^{(1)}_{u_2} = 1, \bar{C}^{(1)}_{u_3} = 2$, in which
\begin{align*}
    u_1 \in U_1:=\Big\{&(3,1,1), (3,1,2), (3,2,1), (3,2,2), \\ &(1,3,1),(1,3,2), (2,3,1), (2,3,2), \\
    &(1,1,3), (1,2,3),(2,1,3),(2,2,3) \Big\}; \\
    u_2 \in U_2:=\Big\{&(1,3,3), (2,3,3), (3,1,3), \\ &(3,2,3), (3,3,1), (3,3,2) \Big\};\\
    u_3 \in U_3:=\Big\{&(3,3,3)\Big\}.
\end{align*}

\end{example}
\begin{remark}
Theorem \ref{theorem:mpot_equivalence_1} requires a simple condition that the sequence $(A_i)$ needs to be monotone increasing. In comparison with \citep{chapel2020partial}, they use the sequence $\big(0, A)$ for the bi-marginal case. That resemblance partly explains why the first equivalence is the more natural extension of the bi-marginal case. The condition  \eqref{asummption:mpot_equivalence_1} could be met in some situations, i.e., the values of $\|\rR_i\|_1$ are the same. However, in general, it is quite restrictive, since the condition requires that the magnitude of those positive measures are close to each other.
\end{remark}
In order to prove the results of both Theorems~\ref{theorem:mpot_equivalence_1} and~\ref{theorem:mpot_equivalence_2}, we need Lemma~\ref{lemma:hypercube} which plays a key role in our proofs. More specifically, this lemma shows multiple ways to transport mass inside the multimarginal plan $\XX$ such that the marginal constraints of $\XX$ still hold. Since Lemma~\ref{lemma:hypercube} works on the hypercube $\mathcal{C}_k = \{0,1\}^k$, we start with some notations and definitions of the hypercube.

\begin{definition}[Hyper-cube and its graph]
Let $\mathcal{C}_k = \{0,1 \}^k = \{(i_1,\ldots,i_k): i_{\ell}\in \{0,1\} \}$ be the hyper-cube of dimension $k$. The graph on hyper-cube $\mathcal{C}_k$ is defined as follow: for $u = (u_1,\ldots,u_k),v = (v_1,\ldots,v_k)\in \mathcal{C}_k$, $u$ and $v$ are connected if and only if there exists only one index $j$ such that $1\leq j\leq k$ and $u_j\neq v_j$. In the language of distance, the Hamming distance between $u$ and $v$ is equal to $1$.
\end{definition}

\begin{definition}[Hyper-rectangle and its graph]
Let $m\geq k$ be two positive integers. Assume that $u = (u_1,\ldots,u_m), v = (v_1,\ldots,v_m)\in \mathbb{Z}^m$ and $\big|\{u_i\neq v_i,1\leq i\leq m\}\big| = k$. The hyper-rectangle which has $u$ and $v$ as two opposite vertices  is defined as follow:
\begin{align*}
\mathcal{P}(u,v)= \big\{ (w_1,\ldots,w_m): w_i \in \{u_i,v_i\}~\forall i \in [m] \big\}.
\end{align*}
Similarly, the graph on the hyper-rectangle is defined as follows : $w = (w_1,\ldots,w_m)$ and $\widetilde{w} = (\widetilde{w}_1,\ldots,\widetilde{w}_m)$ are connected if and only if there exists only one index $\ell$ such that $w_{\ell} \neq \widetilde{w}_{\ell}$.
\end{definition}
In graph theory, the graph of  hyper-rectangle is the same as the graph of the hyper-cube. Hence, the bellow lemma for the hyper-cube could be applied to the hyper-rectangle as well. 

\begin{figure}[!t]
    \centering
    \includegraphics[width=0.6\textwidth]{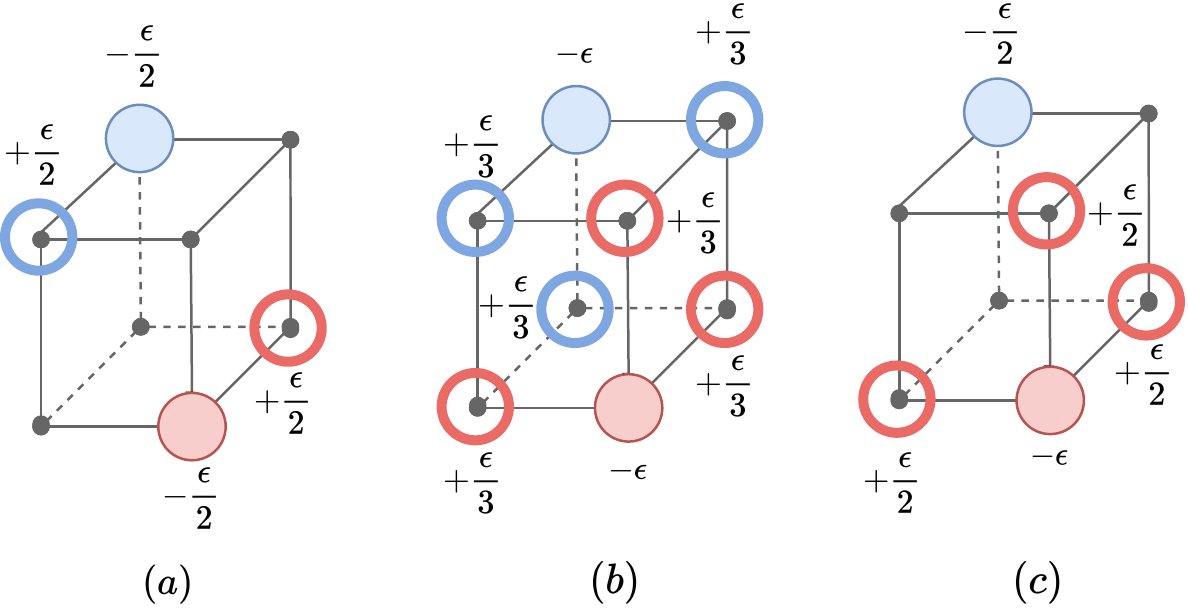}  
    \caption{Visualization of mass-moving procedures in Lemma \ref{lemma:hypercube} for $k=3$. Blue-filled circles represent vertices $\zeros$ and red-filled circles represent vertices $\ones$, while the unfilled circles depict the neighbors of the color-corresponding filled nodes. The number, that is next to the node, i.e., $\epsilon$, is the mass that node gains or loses in the mass-moving procedure.}
    \label{figure:mass_transportation}
\end{figure}

\begin{lemma}[\textbf{Preservative Mass Transportation on Hypercube}]
\label{lemma:hypercube}
Let  $\mathcal{C}_k = \{0,1 \}^k$ be a hyper-cube in the $k$-dimensional space. For each vertex $v$ of the cube  $\mathcal{C}_k$, we assign a mass number $M_v$. Assume that $0 < \epsilon < \min \{ M_{\zeros_k}, M_{\ones_k} \} $, then each of the following operations will keep the quantity 
\begin{align*}
    \sum_{v\in \text{\,one face of $\mathcal{C}_k$}} M_v
\end{align*}
unchanged for any faces of the cube.

(a) We move $\frac{\epsilon}{2}$ mass from vertex $u=\zeros_k$ to its neighbours $\widetilde{u}$ and we move $\frac{\epsilon}{2}$ mass from vertex $v= \ones_k$ to its neighbour $\widetilde{v}$ such that the edge $(u, \widetilde{u})$ is parallel to the edge $(v, \widetilde{v})$.

(b) We move $\frac{\epsilon}{k}$  mass from vertices $\zeros_k$ and $\ones_k$ to each of their neighbours ($\epsilon$ mass in total).

(c) We take away $\epsilon$ and $\frac{\epsilon}{k-1}$ mass from $\ones_k$ and $\zeros_k$, respectively. We add $\frac{\epsilon}{k-1}$ to all $k$ neighbours of $\ones_k$.
\end{lemma}
An illustration of Lemma~\ref{lemma:hypercube} is in Figure~\ref{figure:mass_transportation}, and the proof is deferred to Appendix~\ref{sec:proof_hypercube}.

\begin{figure}[!t]
    \centering
    \includegraphics[width=0.6\textwidth]{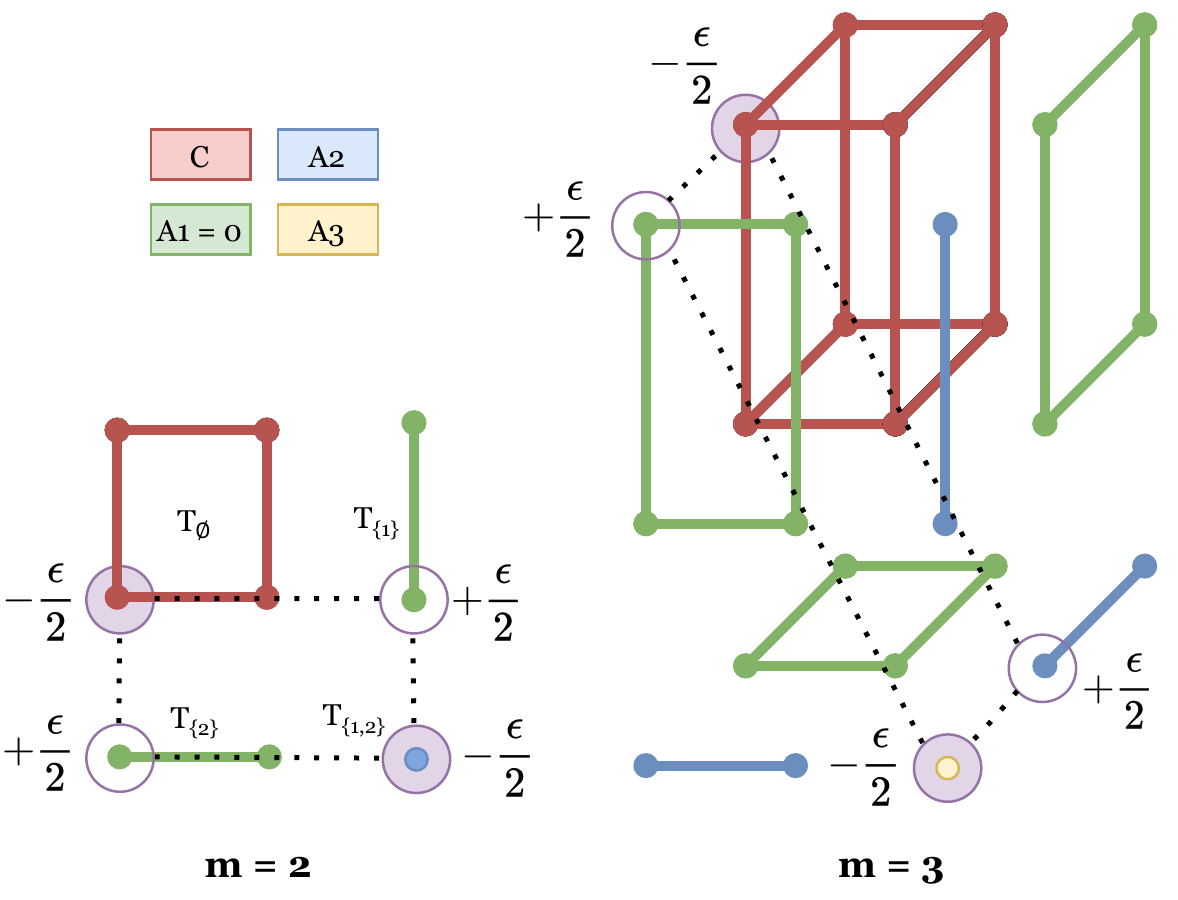}  
    \caption{An illustration for mass-moving procedures in the proof of the first equivalent form in two cases $m = 2$ and $m = 3$, where the red part corresponds to the original supports of the cost plan and others are the extended, $+\frac{\epsilon}{2}$ means adding $\frac{\epsilon}{2}$ mass to that node, and  $-\frac{\epsilon}{2}$ means take away $\frac{\epsilon}{2}$ mass from that node. 
    }
    \label{figure:illustration_equivalence1}
\end{figure}

\begin{proof}[Proof Sketch of Theorem~\ref{theorem:mpot_equivalence_1}]
The full proof of Theorem~\ref{theorem:mpot_equivalence_1} can be found in Appendix~\ref{Sec:proofs}. A key step in the proofs of equivalent forms is to show that $\sum_{v\in T_{\varnothing}}\bar{X}^*_{v}=s$, i.e., the total mass at the portion of $\bar{\XX}^*$ corresponding to the original cost tensor (the zeroth layer) is the total mass we want to transport in the original partial problem. To see this, we start by showing that there is some mass in the zeroth layer, i.e., there exists $u \in T_\varnothing$ such that $\bar{X}^*_{u} > 0$. Next, we prove by contradiction that the mass of $\bar{\XX}^*$ only lies at the zeroth layer and the first layer (which correspond to the red region and the green region in Figure \ref{figure:illustration_equivalence1}, respectively), i.e., $\bar{X}^*_{v}=0$ if $v \notin T_\varnothing \cup \big(\cup_{j=1}^m T_{\{j \}} \big)$. If it is not the case, then we can construct a preservative mass-moving procedure (see Lemma~\ref{lemma:hypercube} and Figure \ref{figure:illustration_equivalence1}) involving the zeroth layer (in particular, the location $u \in T_\varnothing$ with positive mass), and the cost design will invalidate the optimality of $\bar{\XX}^*$. After that, the aforementioned key statement can be attained using some algebraic transformations on marginal constraints. Consequently, we get
\begin{align*}
    \min_{\bar{\XX} \in \Pi(\bar{\rR}_1,\ldots,\bar{\rR}_m)} \langle \bar{\CC}^{(1)}, \bar{\XX}\rangle = \langle \bar{\CC}, \bar{\XX}^* \rangle \geq \min_{\XX \in \Pi^s(\rR_1,\ldots,\rR_m)} \langle \CC, \XX\rangle,
\end{align*}
and we prove that the equality by showing the inverse inequality, which is done via some construction of $\bar{\XX}^*$ based on $\XX^*$.
\end{proof}
\subsection{The Second Equivalent Form}
The first equivalent form requires the total masses $\|\rR_i\|_1$ where $1 \leq i \leq m$ to satisfy the condition~\eqref{asummption:mpot_equivalence_1}, which can be undesirable under general settings of unbalanced measures. To circumvent this limitation of the first equivalence, we propose another equivalent form to the multimarginal POT problem, where the extended cost tensor is more sophisticated and less sparse than that of the first equivalent form. However, no additional conditions are needed for the second equivalence, except the conditions $s\leq \|\rR_j\|_1$ for all $j\in[m]$, which are apparently essential in the original formulation of multimarginal POT. 
\begin{theorem}
\label{theorem:mpot_equivalence_2}
Let $\{D_i\}_{i = 0}^{m}$ be a sequence satisfying $D_0 = \| \CC\|_\infty, D_{m - 1} = 0, D_m > 0$. Furthermore, suppose that $(D_0,D_1,D_2)$ is a concave sequence when $m=3$, while for $m\geq 4$, we assume
\begin{align*}
    \Delta^{(2)}_i \le (m - 1 - i) \Delta^{(2)}_{i + 1} \le 0, \quad \forall i\in[m-3],
\end{align*}
where $\Delta^{(2)}_i := D_{i + 1} + D_{i - 1} - 2 D_{i}$. Let
\begin{align*}
    \bar{\rR}_k^{(2)} = \Big[\rR_k, \sum_{i\neq k} \|\rR_i\|_1 - (m-1) s\Big]
\end{align*}
be extended marginals and $\bar{\CC}^{(2)} = \big[\bar{C}^{(2)}_v: v\in [n+1]^m \big]$, where
\begin{align*}
    \bar{C}_{v}^{(2)} =\begin{cases} C_{v} &\quad v  \in T_\varnothing \\
      D_i & \quad v\in T_S, |S| = i \quad \forall i \in [m]
      \end{cases}
\end{align*}
be an extended cost tensor. Then,
\begin{align}
    \label{eq:mpot_mot_2}
    \min_{\XX \in \Pi^{s}(\rR_{1}, \ldots, \rR_{m})} \langle \CC,\XX\rangle = \min_{\bar{\XX} \in \Pi(\bar{\rR}_{1}^{(2)}, \ldots, \bar{\rR}_{m}^{(2)})} \langle \bar{\CC}^{(2)},\bar{\XX}\rangle.
\end{align}
Similar to Theorem~\ref{theorem:mpot_equivalence_1}, assume that $\Bar{\XX}^*$ is a minimizer of the multimarginal OT problem, then $\XX^*:=\Bar{\XX}^*_{[n]^m}$ is an optimal solution of the multimarginal POT problem.
\end{theorem}
\begin{figure}[!t]
    \centering
    \includegraphics[width=0.6\textwidth]{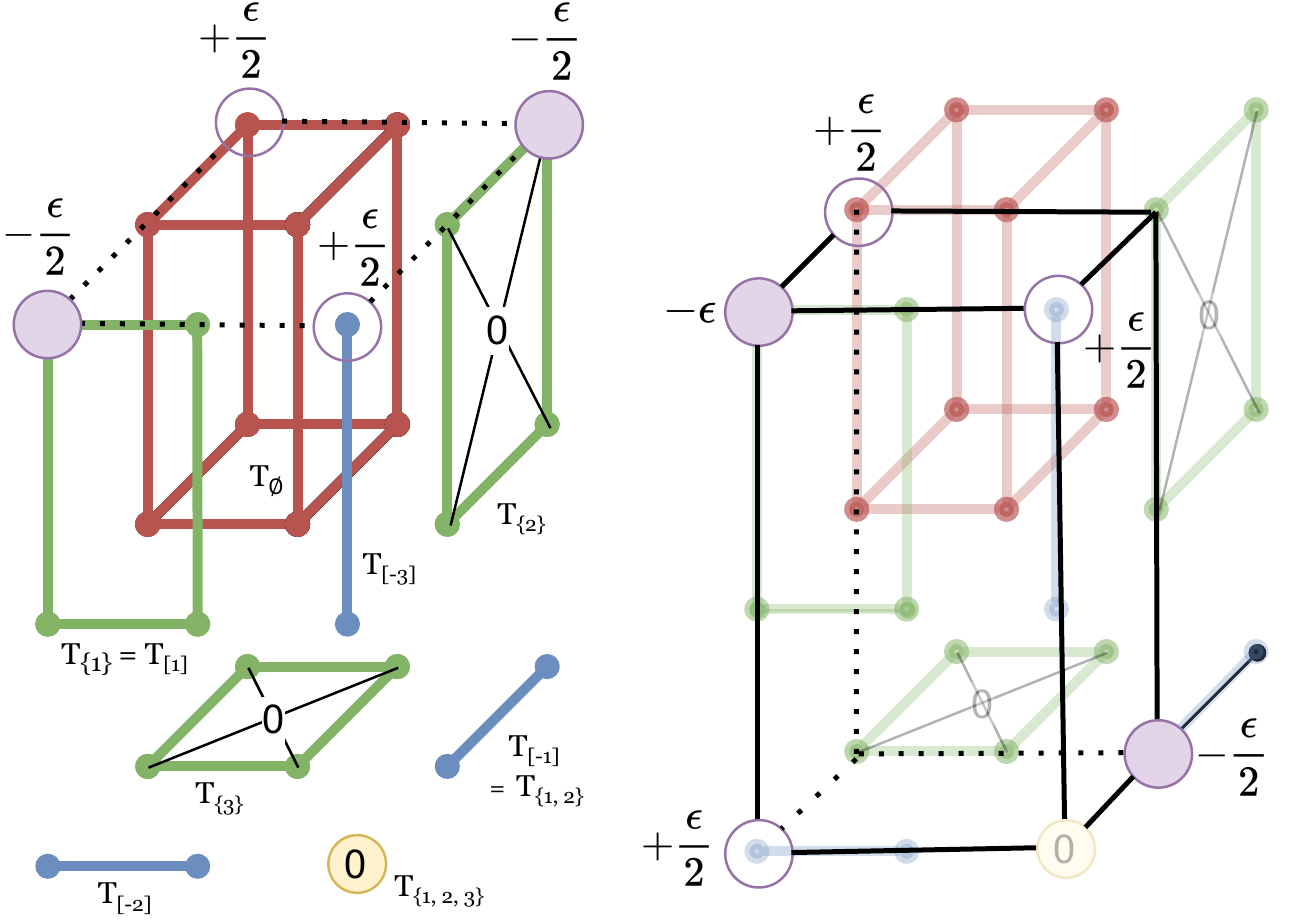}  
    \caption{An illustration for mass-moving procedures in the proof of the second equivalent form when $m = 3$.}
    \label{figure:illustration_equivalence2}
\end{figure}
\begin{proof}[Proof Sketch of Theorem~\ref{theorem:mpot_equivalence_2}]
For the full proof, the readers can refer to Appendix~\ref{sec:proof_theorem:mpot_equivalence_2}. Note that each layer can be decomposed into corresponding sublayers (e.g., the first layer $|S| = 1$ consists of three sublayers $T_{\{1\}}, T_{\{2\}}, T_{\{3\}}$ corresponding to the green region in Figure \ref{figure:illustration_equivalence2}). The first step is to show that the supports of the optimal transportation plan (called \textit{optimal supports}) must lie in a sequence of nested sublayers when their orders are in $[m-2]$; if not, use a mass-moving procedure (depicted on Figure \ref{figure:illustration_equivalence2}, left) built on nodes in two sublayers of the same layer to lower the objective function. This is due to the concavity of the constructed cost sequence $\{D_i\}_{i = 1}^{m - 2}$ (see Theorem~\ref{theorem:mpot_equivalence_2}). The next step is to show that the optimal supports must not lie in the layers of order $1,2,\ldots,m-2$, which relies on the procedure between one of the layers of order $1,\ldots,m-2$ and the  layer of order $m-1$ (see Figure \ref{figure:illustration_equivalence2}, right), and also due to the noted concavity and $D_{m - 1} = 0$. Thus, the optimal supports must lie in  the original layer (i.e., the red $T_{\varnothing}$) and the two last (blue and yellow) layers. Finally, simple derivations show that the total optimal mass on the original layer is $s$, and subsequent reasoning is similar to the last step in the proof of Theorem \ref{theorem:mpot_equivalence_1}.
\end{proof}
\begin{remark}
Theorem \ref{theorem:mpot_equivalence_2} requires some conditions on the second order of difference of the sequence $(D_i)$ that trade off for having no conditions on the $\rR_i$. Given the sequence $(\Delta_i^{(2)})$ and let $\Delta^{(1)}_i:=D_{i+1}-D_{i}$, we could build up the sequence $(D_i)$ as follows:
\begin{align*}
D_i &= \Delta_{i-1}^{(1)} + \ldots + \Delta_{0}^{(1)} + D_0 \\
    &= \Delta_{i-1}^{(2)} + 2 \Delta_{i-2}^{(2)} + \ldots + (i-1) \Delta_1^{(2)} + i\Delta_0^{(1)} + D_0.
\end{align*}
When all the chosen $\Delta_j^{(2)}$ satisfying the conditions $\Delta_j^{(2)} \leq (m-1-j)\Delta_{j+1}^{(2)}\leq 0$ (e.g., $\Delta^{(2)}_{j}=-(m-1-i)!$), we choose $\Delta_0^{(1)} = -\frac{\sum_{i = 1}^{m-2} (m - 1 - i)\Delta_i^{(2)} + D_0}{m-1}$, which leads to the fact that $D_{m-1} = 0$. Then, the sequence $(D_i)$ satisfies all the required conditions. Since the sequence $(D_i)$ is concave, its minimum will be at the end terms of the sequence, which are $D_{m-1}$ and $D_0$. This means that other $D_i$ cannot be negative. 
\end{remark}
\begin{example}
\label{example:second_equivalence}
Similar to Example~\ref{example:first_equivalence}, we illustrate the construction of the extended cost tensor in Theorem~\ref{theorem:mpot_equivalence_2} under the 3-marginal setting, i.e., $m = 3$, and $\|r_{1}\|_{1} = \|r_{2}\|_{1} = \|r_{3}\|_1 = 1$. For simplicity, we set $n=2$, meaning that each marginal has two support points. Based on Theorem~\ref{theorem:mpot_equivalence_2}, the extended cost tensor $\bar{\CC}^{(2)}$ takes the following form
\begin{align*}
    \bar{C}^{(2)}_u = \begin{cases}
        ~D_1, \quad &u\in U_1\\
        ~0, \quad &u\in U_2\\
        ~D_3, \quad &u\in U_3
    \end{cases},
\end{align*}
where we choose $D_1 > D_0 = \max_{u\in [n]^m} C_u$ and $D_3 > 0$ while $U_1, U_2, U_3$ are defined as in Example~\ref{example:first_equivalence}. 
Here, the sequence $D_0,D_1, D_2$ is already concave. 
\end{example}

\subsection{Algorithmic Developments}

\begin{algorithm}[!t]
\caption{\textbf{ApproxMPOT}}
\label{algorithm:entropic_mpot}
\begin{algorithmic}
\STATE \textbf{Parameters:} $\CC, \{
\rR_i\}_{i = 1}^m, s, \eta, \varepsilon$
\STATE Extend the cost tensor $\CC$ into $\bar{\CC}$ and $\{\rR_i\}_{i = 1}^m$ into $\{\bar{\rR}_i\}_{i = 1}^m$ according to Theorem
\ref{theorem:mpot_equivalence_1} or Theorem \ref{theorem:mpot_equivalence_2}
\STATE Compute $\bar{\XX}^k = \mathbf{SinkhornMOT}(\bar{\CC}, \eta, \{\bar{\rR}_i\}_{i = 1}^m, \varepsilon)$, where $k$ is the number of Sinkhorn iterations
\STATE Let $\XX^k = \bar{\XX}^k[1:n, \dots, 1:n]$
\STATE \textbf{return} $\XX^k$
\end{algorithmic}
\end{algorithm}

\noindent In this section, we briefly derive algorithmic procedure to approximate the multimarginal partial optimal transport problem based on the previous equivalences, as well as the computational complexity of the approximating algorithms. Algorithm \ref{algorithm:entropic_mpot} is the approximating algorithm based on the Sinkhorn procedure (namely $\mathbf{SinkhornMOT}$) to solve entropic-regularized multimarginal OT given in \cite[Algorithm 3]{lin2019complexity}. Interestingly, the computational complexity of the multimarginal POT can also be derived from that of multimarginal OT. Before stating that complexity result, we first define the notion of $\varepsilon$-approximated multimarginal partial transportation plan.

\begin{definition}[$\varepsilon$-approximation]
\label{def:epsilon_approx}
The tensor $\widehat{\XX} \in \br_+^{n^{m}}$ is called an \emph{$\varepsilon$-approximated multimarginal partial transportation plan} if $c_k(\widehat{\XX}) \leq r_k$ for any $k \in [m]$ and the following inequality holds true, 
\begin{equation*}
\langle \CC, \widehat{\XX}\rangle \leq \langle \CC, \XX^\star\rangle + \varepsilon,
\end{equation*}
where $\XX^\star$ is defined as an optimal solution of the multimarginal POT problem~\eqref{eq:MPOT_objective}.
\end{definition}
Given Definition~\ref{def:epsilon_approx}, we have the following proposition about the computational complexity of ApproxMPOT algorithm for approximating the multimarginal POT problem.
\begin{proposition}
\label{proposition:entropic_mpot:complexity}
Algorithm \ref{algorithm:entropic_mpot} returns an $\varepsilon$-approximated multimarginal partial transportation plan $X^k$ within $\mathcal{O}\left(\frac{m^{3} (n+1)^{m}\|\bar{\CC}\|_{\infty}^{2} \log (n + 1)}{\varepsilon^{2}}\right)$ arithmetic operations where $\bar{\CC}$ is a given cost tensor in either Theorem~\ref{theorem:mpot_equivalence_1} or Theorem~\ref{theorem:mpot_equivalence_2}.
\end{proposition}
The proof of Proposition~\ref{proposition:entropic_mpot:complexity} is in Appendix~\ref{sec:proof:proposition:entropic_mpot:complexity}.

\section{Empirical Study}
\label{sec:empirical_study}

\begin{figure}[!t]
    \centering
    \includegraphics[width=0.7\textwidth]{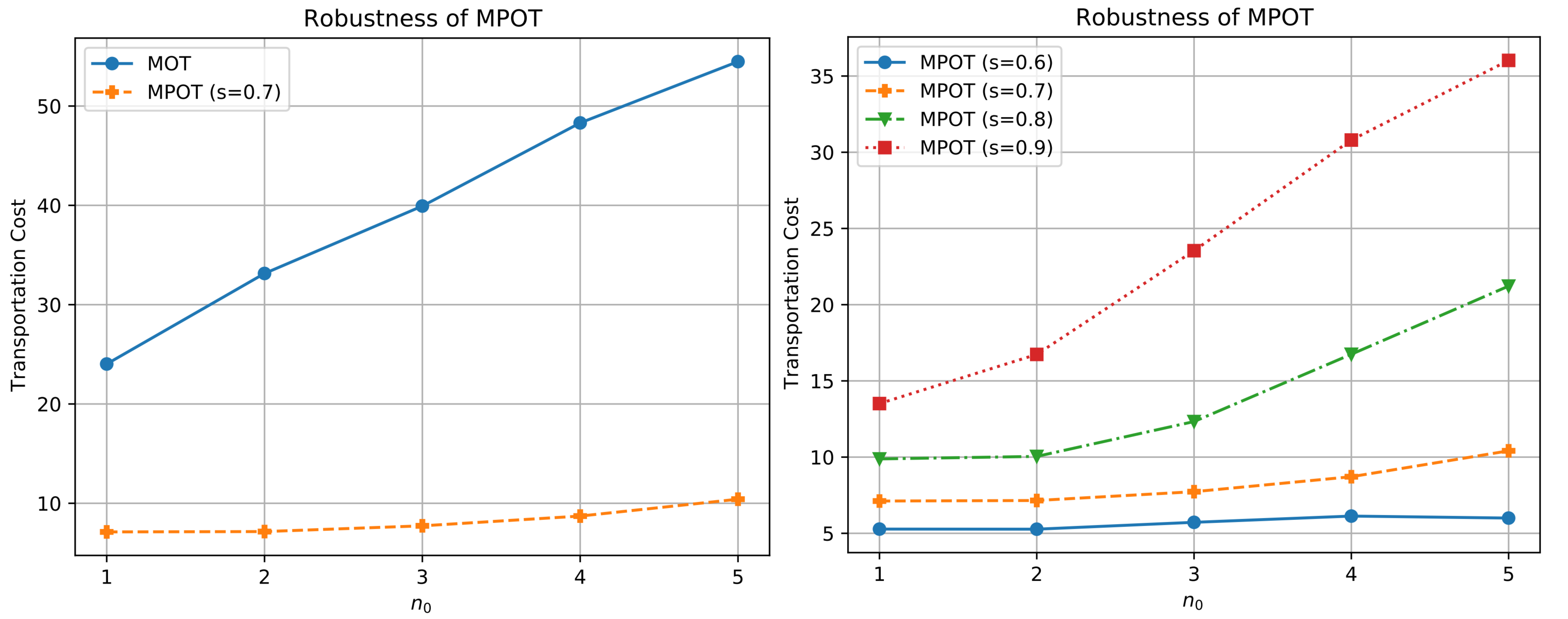}  
    \caption{Robustness of the transportation cost of MPOT compared to MOT when injecting noise into three marginals. \textit{Left}: MPOT and MOT in comparison. \textit{Right}: MPOT with different prespecified mass $s \in \{0.6, 0.7, 0.8, 0.9\}$.}
    \label{figure:robust}
\end{figure}

\paragraph{A Simple Illustration of Robustness.} In this experiment, we empirically verify that MPOT is more robust to outliers compared to MOT, in the sense the optimal MPOT cost is less sensitive in the appearance of noisy marginal support. We consider three empirical measures of 10 supports sampled from $\mathcal{N} \left((0, 0), I_{2} \right)$, $\mathcal{N} \left( (1,1), I_{2} \right)$, and $\mathcal{N} \left( (-1, 1), I_{2} \right)$ respectively, and gradually inject $n_0 \in \{1,2,3,4,5\}$ noisy support points drawn from faraway Gaussians $\mathcal{N} \left( (0,5), I_{2} \right)$, $\mathcal{N} \left( (5,5), I_{2} \right)$, and $\mathcal{N} \left( (-5,5), I_{2} \right)$, respectively. The weights on the total $10+n_0$ support points are set uniformly. The corresponding MOT and MPOT costs (with squared Euclidean ground metric) are reported in Figure~\ref{figure:robust}, which highlights the expected behavior.

\begin{figure}[!t]
    \centering
    \includegraphics[width=0.7\textwidth]{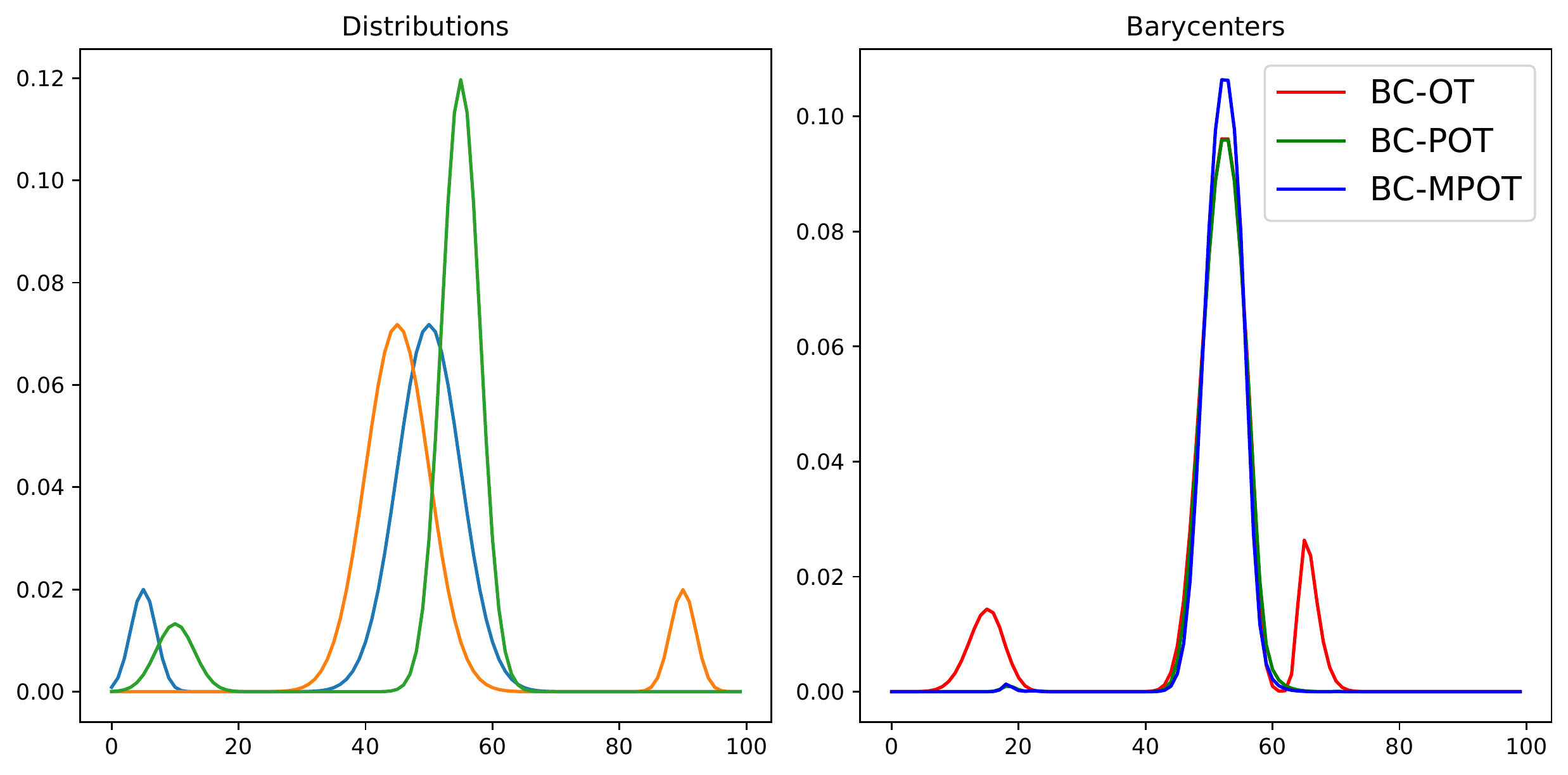}  
    \caption{Robustness of multimarginal partial optimal transport. \textit{Left}: three corrupted Gaussian histograms of $0.9 \cdot \mathcal{N}(50, 25) + 0.1 \cdot \mathcal{N}(5, 4)$ (blue), $0.9 \cdot \mathcal{N}(45) + 0.1 \cdot \mathcal{N}(90, 4)$ (orange), $0.9 \cdot \mathcal{N}(55, 9) + 0.1 \cdot \mathcal{N}(10, 9)$ (green). \textit{Right}: the standard OT barycenter (red), the partial OT barycenter (green), and the barycenter computed by MPOT (blue). The partial algorithms are run with total masses of $0.8$.}
    \label{figure:barycenter_mpot}
\end{figure}

\begin{figure}[!t]
    \centering
    \includegraphics[width=0.6\textwidth]{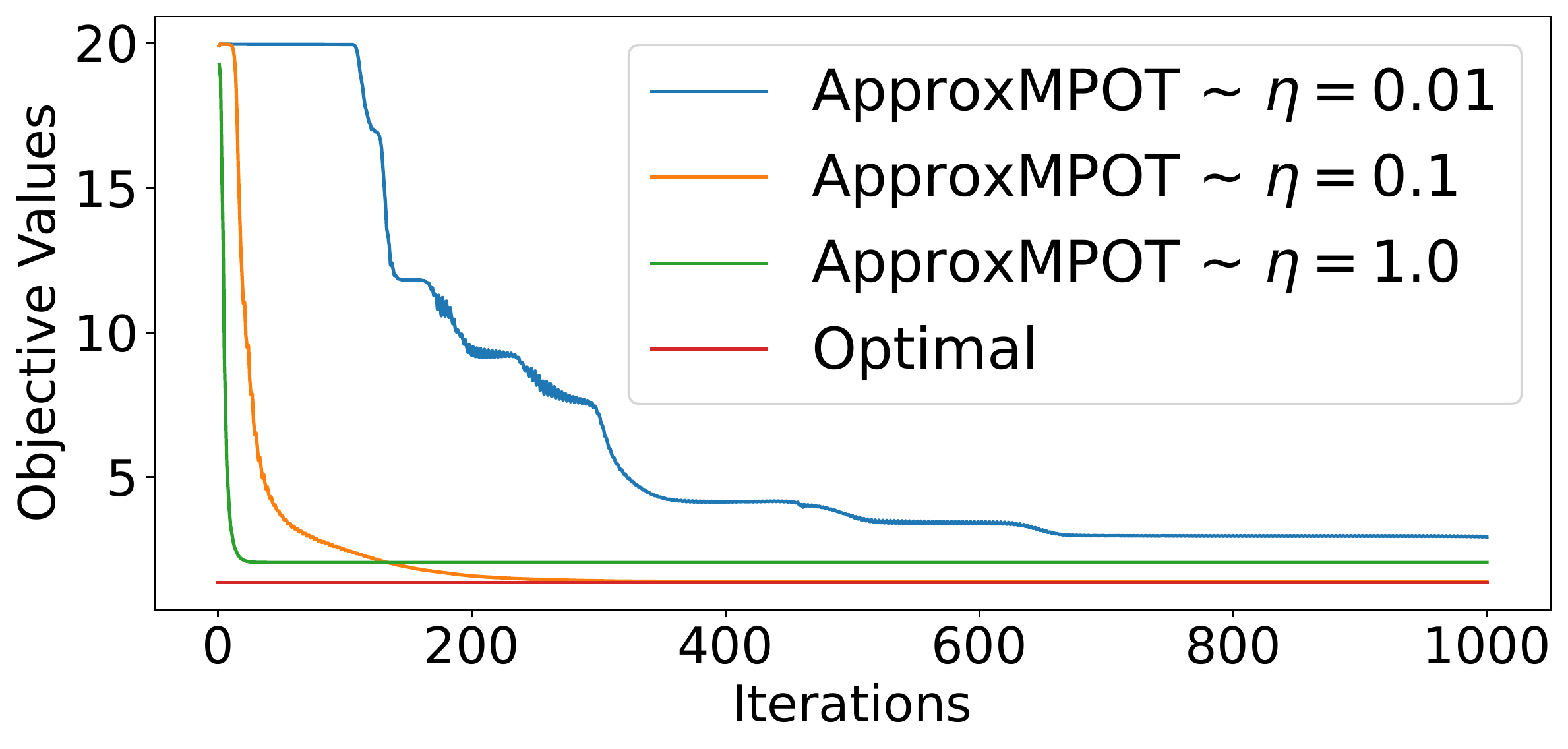}  
    \caption{Convergence of ApproxMPOT. The red line is the optimal value of the unregularized problem, while three other lines correspond to different runs of the algorithm for solving the regularized problems with different values of $\eta$.}
    \label{figure:approx_mpot_convergence}
\end{figure}

\paragraph{Partial Barycenter.} It is known that the multimarginal partial optimal transport is equivalent to the partial barycenter problem \cite[Proposition 1.1]{Kitagawa_MPOT}. In this experiment, we investigate the robust behavior of MPOT by using it to compute the (partial) barycenter of three corrupted Gaussian measures. Specifically, we present each Gaussian measure by a histogram over $100$ support points, where the masses on these points come from a mixture of two Gaussian distribution with weights $0.9$ and $0.1$ (the former corresponds to the true underlying distribution, while the latter is the noise - this simulates Huber's $0.1$-contamination model). The detail can be found in the Figure \ref{figure:barycenter_mpot}, in which we plot three barycenters: one corresponding to the standard optimal transport, one to the partial optimal transport, and one coming from solving a MPOT problem \cite[Section 5.3]{Benamou-2015-Iterative}. These barycenters are computed using convex solvers \citep{cvxpy_rewriting} on the corresponding entropic-regularized formulations with $\eta = 1$ for smoother visualization. It is apparent that the histograms from the partial barycenter problem and from MPOT resemble each other, and are not affected by outliers (while the OT-based histogram is).
\paragraph{Empirical Convergence of ApproxMPOT.} Next, we take a look into the convergence of our approximating algorithm for different regularization values. Specifically, we set $m = 3, n = 10, s = 0.8$ and run the algorithm for $\eta \in \{0.01, 0.1, 1\}$ then plot the objective values over iterations. The optimal value of unregularized problem is computed by a convex solver \citep{cvxpy_rewriting}. As we can see from the plot in Figure \ref{figure:approx_mpot_convergence}, small $\eta$ converges slower, but large $\eta$ may result in an inaccurate approximation.

\section{Conclusion}
\label{Sec:conclusion}
In this paper, we study the multimarginal partial optimal transport problem between $m \geq 2$ (unbalanced) discrete measures with at most $n$ supports. We first establish two equivalent forms of the multimarginal POT problem in terms of the multimarginal OT problem via novel extensions of the cost tensors. Our proof techniques are based on sophisticated procedures of designing extended cost tensors and of moving masses such that the transportation plan will put its mass into appropriate regions. Based on these equivalent forms, we then develop an optimization algorithm, named ApproxMPOT algorithm, for approximating the multimarginal POT with a computational complexity upper bound of the order $\bigOtil(m^3(n+1)^m/ \varepsilon^2)$ where $\varepsilon > 0$ is the tolerance.

\bibliographystyle{abbrvnat}
\bibliography{Arxiv_main}

\appendix
\thispagestyle{empty}

\begin{center}
    {\bf{\LARGE{Supplement to “On Multimarginal Partial Optimal Transport: Equivalent Forms and Computational Complexity”}}}
\end{center}

In this supplementary material, we firstly provide the proof of Theorem~\ref{theorem:mpot_equivalence_1} in Appendix~\ref{Sec:proofs} while the proof of Theorem~\ref{theorem:mpot_equivalence_2} is subsequently presented in Appendix~\ref{sec:proof_theorem:mpot_equivalence_2}. Finally, Appendix~\ref{sec:proof_remaining_results} is devoted to show proofs for Lemma~\ref{lemma:hypercube} and Proposition~\ref{proposition:entropic_mpot:complexity}.

\section{PROOF OF THEOREM~\ref{theorem:mpot_equivalence_1}}
\label{Sec:proofs}
To prove $A=B$, we prove that $A\geq B$ and $B\geq A$, where $A$ and $B$ are the transportation cost of the multimarginal POT and multimarginal OT, respectively. Denote by $\bar{\XX}^*$ the solution of the multimarginal OT problem $\min_{\bar{\XX} \in \Pi(\bar{\rR}_{1}^{(1)}, \ldots, \bar{\rR}_{m}^{(1)})} \langle \bar{\CC}^{(1)},\bar{\XX}\rangle$. We divide the proof of Theorem~\ref{theorem:mpot_equivalence_1} into four key steps.

\noindent
\textbf{Step 1:} We first prove that $\displaystyle \sum_{v \in T_{\varnothing}} \bar{X}^*_v \geq s$. Let $T_{\langle \ell\rangle}:=\{v\in [n+1]^m:v_{\ell}=n+1\}$. For each extended marginal $\bar{\rR}_{\ell}^{(1)}$, considering its last element, we get
\begin{align*}
    \sum_{v\in T_{\langle\ell\rangle}} \bar{X}^*_v &= \frac{1}{m-1}\Sigma_{\rR} - \|\rR_{\ell}\|_1 -\frac{1}{m-1}s.
\end{align*}
Summing over all $\ell \in [m]$ leads to
\begin{align*}
   \sum_{\ell=1}^m \sum_{v\in T_{\langle\ell\rangle}} \bar{X}^*_v 
    = \frac{m}{m-1} \Sigma_{\rR} - \Sigma_{\rR} - \frac{m}{m-1}s  = \frac{1}{m-1}(\Sigma_{\rR} - ms).
\end{align*}
Since $\sum_{v\in \bigcup_{\ell=1}^m T_{\langle\ell\rangle}}\bar{X}^*_v \le \sum_{\ell=1}^m \sum_{v\in T_{\langle\ell\rangle}} \bar{X}^*_v$,
it follows that
\begin{align*}
    \sum_{v\in \bigcup_{\ell=1}^m T_{\langle\ell\rangle}}\bar{X}^*_v \leq \frac{1}{m-1}(\Sigma_{\rR} - ms).
\end{align*}
Moreover, we have $[n]^m= [n+1]^m \backslash \big(\cup_{j=1}^m T_{\langle j\rangle} \big)$. It means that
\begin{align*}
    \sum_{v\in [n]^m} \bar{X}^*_v 
    &= \sum_{v\in [n+1]^m} \bar{X}^*_v - \sum_{v\in \cup_{j=1}^m T_{\langle j\rangle}} \bar{X}^*_v \\
    &\geq \frac{1}{m-1}(\Sigma_{\rR} - s) - \frac{1}{m-1}(\Sigma_{\rR} - ms) \\
    &=s.
\end{align*}
Hence, there exists $u \in [n]^m$ such that $\bar{X}^*_u > 0$.

\noindent
\textbf{Step 2:} We prove that if $\bar{X}_v^* > 0$ then $v\in T_\varnothing \cup \big(\cup_{j=1}^m T_{\{j \}} \big)$. Assume the contrary, which means that there exists $v = (v_1, \dots, v_m)$ in some $T_S$ with $|S| = j > 1$ such that $\bar{X}_v^* > 0$. WLOG, we assume that $v_1= v_2 = n+1$. We build an hyper-rectangle where
\begin{align*}
    u &= (u_1, u_2, u_3,~\ldots, u_m) \in T_\varnothing,\\
    v &= (n+1, n+1, v_3,~\ldots, v_m),
\end{align*}
correspond to $\zeros_k$ and $\ones_k$, respectively. Let $\epsilon = \min \big\{ \bar{X}_u^* , \bar{X}_v^*\big\}$ and apply the procedure in Lemma~\ref{lemma:hypercube}(a) with neighbors
\begin{align*}
    \tilde{u} & = (n+1, u_2, u_3,~\ldots, u_m),\\
    \tilde{v} & = (u_1, n+1, v_3,~\ldots, v_m).
\end{align*}
Then the total cost is changed by
\begin{align*}
    - \frac{\epsilon}{2} (C_u^{(1)} - A_1 + A_j - A_{j-1}),
\end{align*}
which is negative (since $C_u^{(1)} \ge 0, A_1 = 0, A_j \ge A_{j - 1}$), contradictory to the optimality of $\bar{\XX}^*$.

\paragraph{Step 3:} We prove that $\displaystyle \sum_{v\in T_{\varnothing}}\bar{X}_v^* = s$. For each extended marginal $\bar{\rR}_{\ell}^{(1)}$, considering its first $m$-th elements, we get
\begin{align*}
    \sum_{v \in [n + 1]^m \setminus T_{\langle \ell\rangle}} \bar{X}^*_v &= \|\rR_{\ell}\|_1.
\end{align*}
Moreover, the result in Step 2 indicates that $\bar{X}^*_{v} = 0$ for all $v \notin T_\varnothing \cup \big(\cup_{j=1}^m T_{\{j \}} \big)$. Thus,
\begin{align*}
    \sum_{v \in T_\varnothing} \bar{X}^*_v + \sum_{v \in T_{\{j \}}, j \neq \ell} \bar{X}^*_v = \|\rR_{\ell}\|_1.
\end{align*}
Summing over all $\ell \in [m]$ leads to
\begin{align*}
    m \sum_{v \in T_\varnothing} \bar{X}^*_v + (m-1) \sum_{j=1}^m \sum_{v\in T_{\{j\}}} \bar{X}_v^* = \Sigma_{\rR}.
\end{align*}
Note that in Step 1 we have obtained
\begin{align*}
   \sum_{\ell=1}^m \sum_{v\in T_{\{\ell\}}} \bar{X}^*_v = \frac{1}{m-1}(\Sigma_{\rR} - ms).
\end{align*}
From the above equations, we deduce that $\sum_{v\in T_{\varnothing}} \bar{X}_v^* = s$.

\noindent
\textbf{Step 4:} We prove the claimed statement. Let $\bar{\XX}^*_{[n]^m}$ is the sub-tensor of $\bar{\XX}^*$ corresponding to the first $m$ components in each dimension. By construction, $c_k\big(\bar{\XX}^*_{[n]^m}\big) \leq \rR_k$ and from Step 3, $\sum_{v\in [n]^m} (\bar{\XX}^*_{[n]^m})_v = s$. Since for any $v \notin [n]^m$, either the corresponding cost $C_v^{(1)}$ is zero (note that $A_1 = 0$) or the corresponding optimal mass $\bar{X}^*_v$ is zero (see Step 2), we obtain
\begin{align*}
   \langle  \bar{\CC}^{(1)}, \bar{\XX}^* \rangle = \langle \CC, \bar{\XX}^*_{[n]^m} \rangle.
\end{align*}
This means that
\begin{align*}
    \min_{\bar{\XX} \in \Pi(\bar{\rR}_1,\ldots,\bar{\rR}_m)} \langle \bar{\CC}^{(1)}, \bar{\XX}\rangle 
    = \langle \CC, \bar{\XX}^*_{[n]^m} \rangle\geq \min_{\XX \in \Pi^s(\rR_1,\ldots,\rR_m)} \langle \CC, \XX\rangle. 
\end{align*}
The next step is to prove the inverse inequality in order to deduce the equality. Let
\begin{align*}
    \XX^* = \argmin_{\XX\in \Pi^s(\rR_1,\ldots,\rR_m)} \langle \CC, \XX\rangle,    
\end{align*}
we are going to expand $\XX^*$ to become the optimal plan $\widetilde{\XX}^*$ of the MOT problem, where  $\widetilde{X}^*_v = X_v^*$ for all $v \in T_\varnothing$. We also know from Step 2 that $\widetilde{X}^*_v = 0$ for all $v \notin \cup_{j=1}^m T_{\{j \}}$. The problem thus boils down to define $\widetilde{X}^*_v$ for all $v \in \cup_{j=1}^m T_{\{j \}}$ satisfying the  marginal constraints. 

Note that if we view each element of the set $\big\{\widetilde{X}_v^*: v\in T_{\{k\}}\big\}$ as a tensor of $[n]^{m-1}$, then it has $m-1$ marginals which are denoted by $S_{k}^{\ell}$ for $\ell\in [m]$ and $\ell\neq k$. In formula, we have
\begin{align*}
    S_k^{\ell} &= \big(S_{k,1}^{\ell},\ldots, S_{k,n}^{\ell}\big);\\
    S_{k,i}^{\ell} &= \sum_{v: v\in T_{\{k\}}, v_{\ell}= i} \widetilde{X}_v^*,\quad i \in [n].
\end{align*}
The $\ell$-th marginal constraint of $\widetilde{\XX}^*$ is written as
\begin{align*}
(\bar{\rR}_{\ell})_{n+1} &= \sum_{v\in T_{\{\ell\}}} \widetilde{X}^*_v =  \|S_{\ell}^{k}\|_1, \quad k\neq \ell,\\
    c_{\ell}\big(\widetilde{\XX}^*\big) &= \Big(c_{\ell}(\XX^*) + \sum_{k\in [m] \backslash \ell} S_k^{\ell}, (\bar{\rR}_{\ell})_{n+1} \Big).
\end{align*}

Because the $T_{\{k\}}$ are disjoint, we first choose $S_{k}^{\ell}$ satisfying the above equations, and then choose $\big\{\widetilde{X}_v^*: v\in T_{\{k\}} \big\}$ satisfying the marginals $S_{k}^{\ell}$. By simple calculation, we have
\begin{align*}
    \|S_{k}^{\ell}\|_1 = (\rR_{k})_{n+1} &=\frac{1}{m-1}\Sigma_r - \|\rR_k\|_1 - \frac{1}{m-1}s; \\
    \sum_{\ell=1,\ell\neq k}^m \|S_{k}^{\ell}\|_1 &= \|\rR_{\ell} - c_{\ell}(\XX^*)\|_1.
\end{align*}

A trivial construction for $S_k^{\ell}$ is
\begin{align*}
    S_{k}^\ell = \big[\rR_{\ell} - c_{\ell}(\XX^*)\big] \frac{ \|S_k^{\ell}\|_1 }{\|\rR_{\ell} - c_{\ell}(\XX^*)\|_1},
\end{align*}
when $\|\rR_{\ell} - c_{\ell}(\XX^*)\|_1>0$; and $S_k^{\ell} = \mathbf{0}$, when $\|\rR_{\ell} - c_{\ell}(\XX^*)\|_1 = 0$. Given $S_{k}^\ell$, we arbitrarily choose $\widetilde{X}_v^*$ for all $v \in T_{\{\ell\}}$ satisfying their marginals equal $S_k^{\ell}$, i.e.,
\begin{align*}
    \widetilde{X}_{v}^* =  (\rR_k)_{n+1}
        \prod_{\ell \in [m]\backslash k} \frac{S_{k,v_{\ell}}^{\ell}}{\|S_k^{\ell}\|_1} 
\end{align*}
for $v= (v_1,\ldots,v_m)$ and $v_k = n+1$, and $(\rR_k)_{n+1}>0$; otherwise $\widetilde{X}_v = 0$, when $(\rR_k)_{n+1} =0$.

Putting these steps together, we reach the conclusion of the theorem.
\section{PROOF OF THEOREM~\ref{theorem:mpot_equivalence_2}}
\label{sec:proof_theorem:mpot_equivalence_2}
First, we introduce some notations that will be used repeatedly in the proof:
\begin{align*}
    W_{S} &= \sum_{u\in T_S} \bar{X}^*_u \qquad \forall S \subset [m]; \\
    [-i] &= [m] \backslash \{i\} \qquad \forall i \in [m].
\end{align*}

\paragraph{Step 1:} We prove that for $S, S' \subset [m]$ and $1\leq |S|, |S^{\prime}| \leq m-2$, if both $W_S$ and $W_{S^{\prime}}$ are positive then $S \subset S^{\prime}$ or $S^{\prime} \subset S$. 

It is equivalent to show that if $W_S, W_{S^{\prime}} > 0$ then $k:= |S\backslash S^{\prime}| =0$ or $\ell:= |S^{\prime} \backslash S|=0$. Assume the contrary that $k,\ell$ are both positive. Thus, there exist $u \in T_S$ and $v \in T_{S'}$ such that $\bar{X}_u,\bar{X}_v >0$ and using permutation, WLOG we can assume that $u$ and $v$ have the following block forms
\begin{align*}
    u = (u_A, u_B, u_C, u_D), \quad v = (v_A, v_B, v_C, v_D),
\end{align*}
where
\begin{align*}
    &u_A = \underbrace{u_1,\ldots,u_i}_{u_j \neq v_j}, &&v_A = (v_1,\ldots,v_i), \\
    &u_B = (n+1,\ldots,n+1), &&v_B = \underbrace{v_{i+1},\ldots,v_{i+k}}_{\neq n + 1}, \\
    &u_C = \underbrace{u_{i+k+1},\ldots, u_{i+k+\ell}}_{\neq n + 1}, &&v_C = (n+1,\ldots,n+1), \\
    &u_D = u_{i+k+\ell+1}\ldots, u_m, &&v_D = (\underbrace{v_{i+k+\ell +1},\ldots, v_m}_{u_j = v_j}).
\end{align*}

Consider $u$ as $\zeros$ and $v$ as $\ones$ in the cube $\{0, 1\}^{i + k + \ell}$. Since $k, \ell >0$, both $\text{Block B}$ and $\text{Block C}$ are non-empty. For some $b \in [i+1,i+k]$, we set $\tilde{u}^B$ to be neighbour of $u$ such that $\tilde{u}^B_b = v_b$ and $\tilde{v}^B$ to be neighbour of $v$ such that $\tilde{v}^B_b = u_b = n + 1$. For some $c \in [i+k+1,i+k+\ell]$, we construct $\tilde{u}^C$ and $\tilde{v}^C$ similarly. As a result, edges $(u, \tilde{u}^B)$ and $(u, \tilde{u}^C)$ are parallel to edges $(v, \tilde{v}^B)$ and $(v, \tilde{v}^C)$, respectively. Applying the mass-moving procedure in Lemma \ref{lemma:hypercube}(a) twice, the total cost is changed by
\begin{align*}
    \frac{\epsilon}{2} \Big[D_{k+j-1} +  D_{k+j+1} - 2D_{k+j} \Big]+ \frac{\epsilon}{2} \Big[D_{\ell+j -1} + D_{\ell + j +1} - 2D_{\ell+j}\Big].
\end{align*}
which is negative due to the concavity of the sequence $\{D_j\}_{j = 1}^{m - 1}$. Here the condition $|S|,|S^{\prime}| \leq m-2$ guarantees that $k+j+1, \ell+j+1 \leq m-1$. Thus the total cost decreases, contradicted to the optimality of $\bar{\XX}^*$. Hence, $k = 0$ or $\ell = 0$, leading to $S' \subset S'$ or $S \subset S'$. Define 
\begin{align*}
    \mathcal{T} &= \{S: S\in [m], W_S >0\}, \\
    \mathcal{T}^{-} &=\mathcal{T} \backslash \big\{[-1],[-2],\ldots,[-m], [m] \big\}.
\end{align*}
Note that if  $S\in \mathcal{T}^{-}$, then $|S|\leq m-2$. Due to the inclusion property for any two elements  of $\mathcal{T}^{-}$, all elements of $\mathcal{T}^{-}$  could be ordered as a sequence, where the inclusion defines the order in $\mathcal{T}^{-}$. Hence, without loss of generality, we could assume that
\begin{align*}\mathcal{T} \subset \Big\{\varnothing, [1],[2],\ldots, [m-2], [-1],[-2],\ldots,[-m]  ,[m]\Big\}.
\end{align*}

\paragraph{Step 2:} We prove that $W_{[m]} = 0$.

This is equivalent to prove that $\bar{X}^*_{v} = 0$, where $v = (n + 1, \dots, n + 1)$.  Assume the contrary that $\bar{X}^*_{v} > 0$. Assume that there exists $u\in T_S$ for $S\in \big\{ \varnothing, [1],\ldots,[m-2] \big\}$ such that $\bar{X}_u^*>0$.

Applying Lemma \ref{lemma:hypercube}(c) with $u$ as $\zeros$ and $v$ as $\ones$. If $u\in T_{\varnothing}$, then the total cost is changed by
\begin{align*}
    - \frac{\epsilon}{m - 1} C_u - \epsilon \cdot D_m + \frac{\epsilon \cdot m}{m - 1} D_{m - 1},
\end{align*}
which is negative due to $C_u \ge 0$ and $D_m > 0$ and $D_{m - 1} = 0$, contradicted to the optimality of $\bar{\XX}^*$.

If $u\in T_{[i]}$ for $1\leq i\leq m-2$, then the total cost is changed by 
\begin{align*}
    -\frac{\epsilon}{m-i-1} D_{i} - \epsilon \cdot D_m + \epsilon\frac{m-i }{m-i-1}D_{m-1}.
\end{align*}
which is negative due to $D_i > 0$, $D_m >0$ and $D_{m-1}=0$, contradicted to the optimality of $\bar{\XX}^*$.

Hence, either $W_{[m]}=0$ or $W_S = 0$ for all $S\in \big\{\varnothing, [1],\ldots,[m-2] \big\}$. For the second case, from the equations for the marginals of $\bar{\XX}^*$, we have
\begin{align*}
    W_{[-i]} &= \|\rR_i\|_1; \qquad i=1,2,\ldots,m\\
    \sum_{j=1}^m W_{[-j]} - W_{[-i]} + W_{[m]} &= \sum_{j=1}^m \|\rR_j\|_1 - \|\rR_i\|_1  - (m-1)s.
\end{align*}
Adding all the $W_{[-j]}$, we get $\sum_{j=1}^m W_{[-j]} = \sum_{j=1}^m \|\rR_j\|_1$. Comparing with the last equation, we deduce that $W_{[m]} = - (m-1)s$, then $W_{[m]} =0$ for $s\geq 0$. Overall, in all the cases, we obtain $W_{[m]} =0$.

\paragraph{Step 3:} We prove that $W_{\varnothing} = s$.

Recall that  $W_\varnothing = \sum_{v\in [n]^m} \bar{X}_v^*$. Note that at the moment, for all $S \subset [m]$, the quantity $W_S$ is zero except for $W_{[i]}$ where $i \in [m - 2]$ and $W_{[-i]}$ where $i \in [m]$. Thus, considering only these non-zero terms,  we  can deploy the system of equations from the marginals as follows:
\begin{itemize}[leftmargin=7mm]
    \item[(1)] For the extended marginal $\bar{\rR}_1^{(2)}$,
    \begin{itemize}[leftmargin=10mm]
        \item[(1.1)] for the first $n$ elements, which corresponds to coordinates $u$ satisfying $u_{1} \neq n + 1$, the equation is
        \begin{align*}
            W_\varnothing + W_{[-1]} = \|\rR_1\|_1;
        \end{align*}
        \item[(1.2)] for the last element, which corresponds to coordinates $u$ satisfying $u_{1} = n + 1$, the equation is
        \begin{align*}
            \sum_{i = 1}^{m - 2} W_{[i]} + \sum_{i = 2}^{m} W_{[-i]} = \sum_{i = 2}^m \|\rR_i\|_1 - (m-1)s. 
        \end{align*}
    \end{itemize}

    \item[(2)] For the extended marginals $\bar{\rR}_i^{(2)}$ (where $i \in \{2,\ldots,m\}$), which corresponds to coordinates $u= (u_1,\ldots,u_m)$ satisfying $u_{i} \neq n + 1$, the equations are
    \begin{align*}
        W_\varnothing + \sum_{j=1}^{i - 1} W_{[j]} + W_{[-i]} &= \|\rR_i\|_1, ~2 \le i \le m - 2, \\
         W_\varnothing + \sum_{j=1}^{m-2} W_{[i]} + W_{[-i]} &= \|\rR_{m-1}\|_1, ~i \in \{m - 1, m\} .
    \end{align*}        
\end{itemize}
Taking the sum of the last $m-1$ equations for $i = 2,\ldots,m$, we obtain
\begin{align*}
(m-1)W_\varnothing + \sum_{i=1}^{m-2} (m-i) W_{[i]} + \sum_{i =2}^m W_{[-i]} &= \sum_{i=2}^m \|\rR_i\|_1
\end{align*}
Compare it with the second equation, 
\begin{align*}
    \sum_{i=1}^{m-2} W_{[i]} + \sum_{i=2}^{m} W_{[-i]} + (m-1)s= \sum_{i=2}^{m} \|\rR_i\|_1,
\end{align*}
we obtain that
\begin{align*}
    (m-1) W_\varnothing + \sum_{i=1}^{m-2} (m - i - 1) W_{[i]}  = (m-1) s.
\end{align*}
There are two scenarios:
\begin{itemize}
    \item $W_\varnothing < s. \quad$Since $\|\rR_1\|_1 \geq s$, we have $\|\rR_1\|_1 > W_\varnothing$. Recall that $W_\varnothing + W_{[-1]} = \|\rR_1\|_1$, hence $W_{[-1]} > 0$. Furthermore, $W_\varnothing < s$ implies that there exists $i \in [m - 2]$ that $W_{[i]} > 0$.
    \item $W_\varnothing = s. \quad$Since $W_{[i]} \ge 0$ for all $i \in [m - 2]$, we have $W_\varnothing = s$ and $W_{[i]} = 0$ for all $i \in [m - 2]$.
\end{itemize}
The second scenario is exactly what we want to prove. Thus, we will show that the first scenario is impossible.

For $S = [i]$ and $S' = [-1]$, we consider some $u \in T_{S}$ as $\ones_k$ and some $v \in T_{S'}$ as $\zeros_k$ in the cube $\{0, 1\}^{k}$ (where $k = m - i + 1)$, which have the forms
\begin{align*}
    u = ( &n + 1, \underbrace{n + 1, \dots, n + 1}_{i - 1 \text{ times}}, \underbrace{u_{i + 1}, \dots, u_m}_{\neq n + 1}) \\
    v = ( &\underbrace{v_1}_{\neq n + 1}, \underbrace{n + 1, \dots, n + 1}_{i - 1 \text{ times}}, n + 1, \dots, n + 1)
\end{align*}
Since $W_S, W_{S'} > 0$, there exists $u\in T_S$ and $v\in T_{S^{\prime}}$ such that $\bar{X}^*_{u}, \bar{X}^*_{v} > 0$. Applying the procedure in Lemma \ref{lemma:hypercube}(c) and noting that $D_{m - 1} = 0$, the total cost is changed at least  by
\begin{align*}
    - \epsilon \cdot D_i + \frac{\epsilon}{k - 1}[D_{i-1} + (k - 1)D_{i + 1}] = \frac{\epsilon}{k - 1} [D_{i - 1} + (k - 1)D_{i+1} - (k - 1) D_i].
\end{align*}
Let $\Delta^{(1)}_j := D_{j+1}- D_j$ for $0 \le j \le m - 2$ that leads to  $\Delta_j^{(2)} := \Delta_j^{(1)} - \Delta_{j-1}^{(1)}$ for $1 \le j \le m - 2$. Note that $k = m - i + 1$ where $i \in [m - 2]$ and $D_{m - 1} = 0$, the term inside the above bracket can be written as
\begin{align*}
    (m-i) \Delta_i^{(1)} +  (D_{i - 1} - D_{m - 1}) & \\
    =~&(m-i) \Delta_i^{(1)} - \big[\Delta_{i-1}^{(1)} + \Delta_i^{(i)} + \ldots + \Delta_{m-2}^{(1)}\big] \\
    =~&\Delta_i^{(2)} - \sum_{j=i+1}^{m-2} \big[\Delta_{i+1}^{(2)} + \ldots + \Delta_j^{(2)} \big] \\
    =~&\Delta_i^{(2)} - \sum_{j = i+1}^{m - 2} (m - 1 - j) \Delta_j^{(2)} \\
    =~&[\Delta_i^{(2)} - (m-1-i) \Delta_{i+1}^{(2)} + \Delta_{i+1}^{(2)}] - \sum_{j=i+2}^{m-2} (m-1-j) \Delta_j^{(2)} \\
    \leq~&\Delta_{i+1}^{(2)} - \sum_{j=i+2}^{m-2} (m-1-j) \Delta_j^{(2)} \\
    \leq~&\ldots \leq \Delta_{m-2}^{(2)} \leq 0,
\end{align*}
where the inequalities in the last  two lines come from the condition that
$\Delta_j^{(2)}\leq (m-1-j) \Delta_{j+1}^{(2)}$.

Thus, we have proved that
\begin{align*}
    \min_{\bar{\XX}\in \Pi(\bar{\rR}_1,\ldots, \bar{\rR}_m)}\langle \bar{\CC}^{(2)}, \bar{\XX}\rangle = \langle \CC, \bar{\XX}_{[n]^m}^*\rangle  \geq \min_{\XX\in \Pi^s(\rR_1,\ldots,\rR_m)} \langle \CC,\XX\rangle.
\end{align*}

\paragraph{Step 4:} We prove the claimed statement.
The next step is to prove the inverse inequality in order to deduce the equality.

For a $\XX^*$ such that $c_k(\XX^*) \leq \rR_k$ and  $\sum_{v\in [n]^m} X_v^* = s$. Then transport tensor map $\widetilde{\XX}^*$ could be obtained from expanding the tensor matrix $\XX^*$ as follows: For $u = (n+1,\ldots,n+1, i_k, n+t,\ldots,n+1) \in T_{[-k]}$ and $1\leq i_k \leq n$
\begin{align*}
    \widetilde{X}_{u}^* = (\rR_k)_{i_k} - \sum_{v\in [n]^m, v_k = i_k} X_v^*.
\end{align*}
For other $u\notin [n]^m\bigcup_{k=1}^m T_{[-k]}$, we set $\widetilde{X}_u^* = 0$. Hence, the  $\widetilde{\XX}^*$ satisfies the marginal constraints. It means that
\begin{align*}
    \langle \CC, \XX^* \rangle  = \langle \bar{\CC}^{(2)}, \widetilde{\XX}^*\rangle \geq \min_{\XX \in \Pi(\bar{\rR_1},\ldots,\bar{\rR}_m)} \langle \bar{\CC}^{(2)},\XX \rangle.
\end{align*}
Putting these steps together, we obtain the conclusion of the theorem.
\section{PROOF OF REMAINING RESULTS}
\label{sec:proof_remaining_results}
\subsection{Proof of Lemma~\ref{lemma:hypercube}}
\label{sec:proof_hypercube}
First, let $F$ be the face that we take the sum. Note that $F$ consists of vertices in the form $(i_1,\ldots,i_k)$ where one of the  $i_{\ell}$ is  constant. 

\noindent (a)  Let $F$ be a face of the cube, WLOG we have 
\begin{align*}
F = \big\{(0,i_2,\ldots,i_m)  \big\}
\end{align*}
\paragraph{Case 1.} $\widetilde{u} = (0,1,0,\ldots,0)$, $F$ loses $\frac{1}{2}\epsilon$ mass at $(0,\ldots,0)$ and gains $\frac{1}{2}\epsilon$ mass at $(0,1,0,\ldots,0)$. So the total mass on $F$ are unchanged.

\paragraph{Case 2.} $\widetilde{u} = (1,0,\ldots,0)$, $F$ loses $\frac{1}{2}\epsilon$ mass at $(0,\ldots,0)$ and gains $\frac{1}{2}\epsilon$ mass at $(0,1,\ldots,1)$, since the edge between $(0,\ldots,0)$ and $(1,0,\ldots,0)$ is parallel to the edge between $(1,,\ldots,1)$ and $(0,1,\ldots,1)$. \\

\noindent (b) Note that if we do it multiple time of moving masses from $u$ and $v$ to theirs neighbours with different $\delta = \frac{1}{k} \epsilon$ mass, the property of unchanged mass sum on a given face still holds. \\

\noindent (c) Due to the symmetric of the transportation for case (c), we could assume that $\ell = 1$. If $i_1 = 0$, then the face $F$ contains all vertices of $(0,i_2,\ldots,i_m)$. Apparently, $F$ contains $(0,\ldots,0)$ and only one neighbour of $\ones_k$ which is $(0,1,\ldots,1)$. Hence the quantity $\sum_{v\in F}M_v$ is unchanged.

If $i_1 = 1$, the face $F$ contains  all vertices of $(1,i_2,\ldots,i_m)$. Here, $F$ contains vertex $\ones_k$ and $m-1$ vertices which are neighbours of $(1,,\ldots,1)$, $F$ does not contain $(0,1,\ldots,1)$ which is also neighbour of $(1,1,\ldots,1)$. Overall, the total weight $M_v$ of the face $F$ are unchanged, since $-\epsilon + (m-1) \frac{\epsilon}{m-1} = 0$.

As a consequence, we reach the conclusion of the lemma.
\subsection{Proof of Proposition~\ref{proposition:entropic_mpot:complexity}}
\label{sec:proof:proposition:entropic_mpot:complexity}
From \cite[Theorem 4.5]{lin2019complexity}, we have
\begin{align*}
    \langle \bar{\CC}, \bar{\XX}^k \rangle - \langle \bar{\CC}, \bar{\XX}^* \rangle \le \varepsilon
\end{align*}
within $O\left(\frac{m^{3} (n+1)^{m}\|C\|_{\infty}^{2} \log (n + 1)}{\varepsilon^{2}}\right)$ arithmetic operations where $\bar{\CC}$ is a cost tensor in either Theorem~\ref{theorem:mpot_equivalence_1} or Theorem~\ref{theorem:mpot_equivalence_2}. From Theorem \ref{theorem:mpot_equivalence_1} and Theorem \ref{theorem:mpot_equivalence_2}, $\langle \CC, \XX^* \rangle = \langle \bar{\CC}, \bar{\XX}^* \rangle$. Moreover, since $\CC$ is a sub-tensor of the non-negative tensor $\bar{\CC}$ by construction, and $\bar{\XX}^k$ is also non-negative, we get $\langle \CC, \XX^k \rangle \le \langle \bar{\CC}, \bar{\XX}^k \rangle$. Hence, $\langle \CC, \XX^k \rangle - \langle \CC, \XX^* \rangle \le \langle \bar{\CC}, \bar{\XX}^k \rangle - \langle \bar{\CC}, \bar{\XX}^* \rangle \le \varepsilon$, completing the proof.

\end{document}